\pdfoutput=1

\documentclass{article}

\usepackage[final, nonatbib]{neurips_2019}

\usepackage[utf8]{inputenc} 
\usepackage[T1]{fontenc}    
\usepackage{hyperref}       
\usepackage{url}            
\usepackage{booktabs}       
\usepackage{amsfonts}       
\usepackage{nicefrac}       
\usepackage{microtype}      
\usepackage{todonotes}

\usepackage{amssymb,amsmath,amsthm,graphicx,color,algorithmic,enumerate,algorithm,subcaption,float,multirow,appendix,verbatim}
\usepackage[export]{adjustbox}  
\usepackage[capitalize]{cleveref}

\newtheorem{theorem}{Theorem}
\newtheorem{corollary}[theorem]{Corollary}
\newtheorem{lemma}[theorem]{Lemma}
\newtheorem{proposition}[theorem]{Proposition}
\newtheorem{axiom}[theorem]{Axiom}
\newtheorem{definition}[theorem]{Definition}
\newtheorem{remark}[theorem]{Remark}

\newcommand{\iali}[1]{\begin{align}#1\end{align}}

\newcommand{\be}{\begin{enumerate}}
\newcommand{\ee}{\end{enumerate}}
\newcommand{\bi}{\begin{itemize}}
\newcommand{\ei}{\end{itemize}}
\newcommand{\ba}{\begin{array}}
\newcommand{\ea}{\end{array}}
\newcommand{\bt}{\begin{tabular}}
\newcommand{\et}{\end{tabular}}
\newcommand{\btb}{\begin{tabbing}}
\newcommand{\etb}{\end{tabbing}}
\newcommand{\bfg}{\begin{figure}}
\newcommand{\efg}{\end{figure}}
\newcommand{\bsl}{\begin{slide}}
\newcommand{\esl}{\end{slide}}
\newcommand{\bthm}{\begin{theorem}}
\newcommand{\ethm}{\end{theorem}}
\newcommand{\bcor}{\begin{corollary}}
\newcommand{\ecor}{\end{corollary}}
\newcommand{\blem}{\begin{lemma}}
\newcommand{\elem}{\end{lemma}}
\newcommand{\bprop}{\begin{proposition}}
\newcommand{\eprop}{\end{proposition}}
\newcommand{\basm}{\begin{assumption}}
\newcommand{\easm}{\end{assumption}}
\newcommand{\baxm}{\begin{axiom}}
\newcommand{\eaxm}{\end{axiom}}
\newcommand{\bdfn}{\begin{definition}}
\newcommand{\edfn}{\end{definition}}
\newcommand{\brmk}{\begin{remark}}
\newcommand{\ermk}{\end{remark}}
\newcommand{\balg}{\begin{algorithm}}
\newcommand{\ealg}{\end{algorithm}}
\newcommand{\bntn}{\begin{notation}}
\newcommand{\entn}{\end{notation}}
\newcommand{\bexm}{\begin{example}}
\newcommand{\eexm}{\end{example}}
\newcommand{\bpf}{\begin{proof}}
\newcommand{\epf}{\end{proof}}

\DeclareMathOperator{\real}{Re}

\DeclareMathOperator{\id}{id}

\DeclareMathOperator{\KL}{KL}
\DeclareMathOperator{\MI}{MI}

\renewcommand{\d}{\text{d}}

\newcommand{\bR}{\mathbb{R}}

\newcommand{\bN}{\mathbb{N}}

\newcommand{\cN}{\mathcal{N}}

\newcommand{\cC}{\mathcal{C}}
\newcommand{\cS}{\mathcal{S}}

\newcommand{\cM}{\mathcal{M}}

\newcommand{\cX}{\mathcal{X}}

\newcommand{\cZ}{\mathcal{Z}}
\newcommand{\cW}{\mathcal{W}}

\newcommand{\norm}[1]{\left\|#1\right\|}

\newcommand{\kld}[2]{\KL\left(#1\,\middle|\,#2\right)}
\newcommand{\eps}{\varepsilon}

\graphicspath{{figures/}{code_fig1/}}

\title{Informative GANs via Structured\\ Regularization of Optimal Transport}

\author{%
  Pierre Br{\'e}chet\\
  Technical University of Munich\\
  \texttt{brechet@in.tum.de}\\
  \And
  Tao Wu\\
  Technical University of Munich\\
  \texttt{tao.wu@tum.de}
  \And
  Thomas M\"ollenhoff\\
  Technical University of Munich\\
  \texttt{thomas.moellenhoff@tum.de}
  \And
  Daniel Cremers\\
  Technical University of Munich\\
  \texttt{cremers@tum.de}
}

\begin{document}

\maketitle

\begin{abstract}
  We tackle the challenge of disentangled representation learning in generative adversarial networks (GANs) from the perspective of regularized optimal transport (OT). Specifically, a smoothed OT loss gives rise to an implicit transportation plan between the latent space and the data space. Based on this theoretical observation, we exploit a structured regularization on the transportation plan to encourage a prescribed latent subspace to be informative. This yields the formulation of a novel informative OT-based GAN. By convex duality, we obtain the equivalent view that this leads to perturbed ground costs favoring sparsity in the informative latent dimensions. Practically, we devise a stable training algorithm for the proposed informative GAN. Our experiments support the hypothesis that such regularizations effectively yield the discovery of disentangled and interpretable latent representations. Our work showcases potential power of a regularized OT framework in the context of generative modeling through its access to the transport plan. Further challenges are addressed in this line.
  
\end{abstract}

\section{Introduction} \label{sec:intro}
A central challenge in machine learning is that of \emph{disentangled representation learning} \cite{bengio2013representation} where the goal is to infer an interpretable 
and low-dimensional latent representation of the data. In this paper, we consider the setting of generative adversarial networks (GANs) \cite{Goo+14} 
which aim to fit the data distribution $\nu$ as the transformation of some low-dimensional latent distribution $\zeta$ by a generator map $G$.
Ideally, in a disentangled setting, variations of the latent code $z \sim \zeta$ should correspond to interpretable changes in the output $G(z)$. 

Unlike existing works for disentangled and informative learning of GANs \cite{CDHSSA16}, our aim is to tackle this problem naturally within the framework of entropically smoothed optimal transport. 

Optimal transport \cite{Vil08,San15,PeCu19} and smoothed formulations thereof \cite{Cut13,GPC18,FSVATP19} have emerged as a promising and flexible framework for density fitting and generative modeling. In this vein, Sinkhorn losses were introduced in order to avoid biases due to the smoothing \cite{GPC18} and later put onto firm theoretical grounds \cite{GCBCP19,FSVATP19}. They were further applied to generative modeling and shown capable of delivering promising performance \cite{SBRL18}. 

One interesting aspect of smoothed OT formulations we would like to focus on is the availability of the transportation plan via primal-dual optimality conditions. 
We will consider an implicit transportation plan which is given by a joint distribution over latent and visible variables. If this distribution
were to coincide or be close to the true (unknown) joint distribution, it would certainly be useful for downstream tasks beyond generative modeling such as classification, uncertainty estimation or reinforcement learning.

As argued by the recent work \cite{locatello2018challenging}, 
without inductive biases or regularizations there is no a priori reason why the learned transportation plan should be close to a ``true'' joint distribution with semantically meaningful latent variables. 
In this work, we encourage such inductive biases by adding an additional structured (concave) entropic regularization term
to the formulation. The idea is to minimize the entropy of the transportation plan after marginalizing out an uninformative subset of the latent variables,
thereby making the remaining variables more informative. 

The other important aspect is the ground cost, which ideally is a semantically meaningful metric on the data space. To improve upon simple choices 
such as $\ell_2$-distances on pixel level, the works \cite{GPC18,SBRL18,li2017mmd} 
learn the cost in an adversarial fashion. We show that a dual interpretation of our proposed structured regularization term connects to learned costs, 
but with a bias towards making transport on the informative dimensions cheaper.
Due to simplicity, we will fix a simple $\ell_1$ or squared $\ell_2$ cost, 
which will be however be reasonable for the considered applications. 

We organize the rest of the paper as follows:
\begin{itemize}
  \item In Section~\ref{sec:ot} we review necessary background on (smoothed) optimal transport and generative modeling. Section~\ref{sec:sink} motivates the use of the Sinkhorn loss, confirming numerically that it helps alleviate variance collapse due to smoothing of the OT loss.

  \item In Section~\ref{sec:htp}, we expose the implicit transportation plan arising from the smoothed OT loss. Under the principle of mutual information maximization, this leads us to an informative generative model via structured entropic regularization on this plan; see Section~\ref{sec:infosink}. By convex duality, we offer an interpretation as a (biased) learning of the ground cost. We further devise a practical training algorithm for deep generative models in Section \ref{sec:train}. 

  \item Our experiments on MNIST in Section~\ref{sec:exp} show that the proposed structured regularization indeed encourages disentangled latent representations. Section~\ref{sec:disc} concludes the paper with a discussion.

\end{itemize}

\begin{figure*}[t!]
   \centering
\begin{subfigure}[t]{0.245\textwidth}
   \text{\small$\cW_{c,\eps}:c(x,y) = \|x-y\|_1$}
   \includegraphics[width=\linewidth,max height=1.00\textheight]{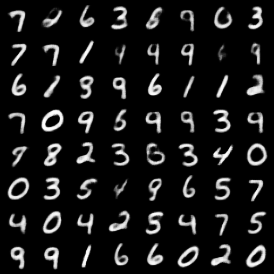}
\caption{}
   \label{fig:debias_a}
\end{subfigure}
\begin{subfigure}[t]{0.245\textwidth}
    \text{\small $\cS_{c,\eps}: c(x,y) = \norm{x-y}_1$.}
   \includegraphics[width=\linewidth,max height=1.00\textheight]{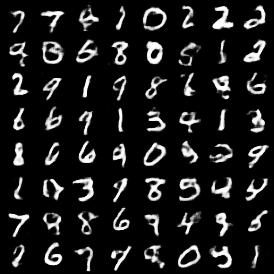}
    \caption{}
   \label{fig:debias_c}
\end{subfigure}
\begin{subfigure}[t]{0.245\textwidth}
    \captionsetup{justification=centering}
    \text{\small $\cW_{c,\eps}: c(x,y)= \norm{x-y}_2^2$.}
   \includegraphics[width=\linewidth,max height=1.00\textheight]{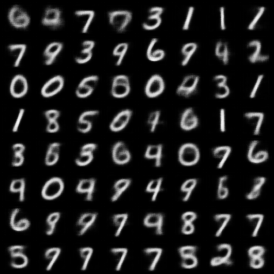}
   \caption{}
   \label{fig:debias_b}
\end{subfigure}
\begin{subfigure}[t]{0.245\textwidth}
    \captionsetup{justification=centering}
    \text{\small $\cS_{c,\eps}: c(x,y)= \norm{x-y}_2^2$.}
   \includegraphics[width=\linewidth,max height=1.00\textheight]{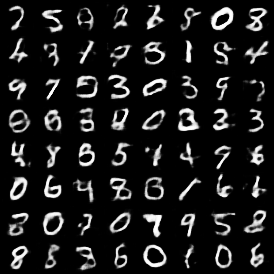}
   \caption{}
   \label{fig:debias_d}
\end{subfigure}
   \caption{Comparison of smoothed OT loss $\cW_{c,\eps}$ and Sinkhorn loss $\cS_{c,\eps}$, all with $\varepsilon = 0.001$.
   The entropic bias carried by the smoothed OT loss causes the generated samples to collapse on
   median (\subref{fig:debias_a}) and mean (\subref{fig:debias_b}) of clusters in the data.
   As shown in (\subref{fig:debias_c}) and (\subref{fig:debias_d}), the collapse is fixed
   by the debiasing term in the Sinkhorn loss.}
   \label{fig:debias}
\end{figure*}
\section{Generative Modeling with the Sinkhorn Loss}
In this section, we review optimal transport in the context of generative modeling, and then draw attention to the recently emerged Sinkhorn loss arising from optimal transport.
For the rest of this paper, we adopt the measure-theoretic notations from \cite{PeCu19}.

\subsection{Background on optimal transport and generative modeling}
\label{sec:ot}
Let $c\in\cC(\cX\times\cX;\bR_+)$ be the ground cost over the compact metric space $\cX$
such that $c(x,y)=0$ for all $x=y$.
The optimal transport (OT) loss, traced back to Kantorovich \cite{Kan42}, between two probability measures $\mu,\nu\in\cM_+^1(\cX)$ is defined as
\iali{
\cW_{c}(\mu,\nu) &= \inf_{\gamma\in\Gamma(\mu,\nu)} \int_{\cX\times\cX}c(x,y)\d\gamma(x,y), \label{eq:wd}\\
\Gamma(\mu,\nu) &= \left\{\gamma\in\cM_+^1(\cX\times\cX): \pi_{x\sharp}\gamma=\mu,\pi_{y\sharp}\gamma=\nu \right\}. \label{eq:wd2}
}
Given $G:\cZ\to\cX$ and $\zeta\in\cM_+^1(\cZ)$, $G_\sharp\zeta$ denotes the pushforward measure \cite[Definition 2.1]{PeCu19}:
\iali{
\forall h\in\cC(\cX;\bR): \int_\cX h(x) d G_\sharp\zeta(x) = \int_\cZ h(G(z))\d\zeta(z).
}
With $\pi_x(x,y)=x$ and $\pi_y(x,y)=y$, the notations of pushforward measures $\pi_{x\sharp}\gamma$ and $\pi_{y\sharp}\gamma$ in \eqref{eq:wd2} agree with the marginalization operations. The optimization variable $\gamma$ is referred to as the transportation plan.

In generative modeling, one aims to learn a generator $G\in\cC(\cZ;\cX)$ that generates a realistic sample $x=G(z)$ from a latent code $z\sim\zeta$ under a fixed distribution $\zeta\in\cM_+^1(\cZ)$.
Equivalently speaking, one seeks for $G$ such that the pushforward measure $G_\sharp\zeta$ is as close as possible to a given empirical measure $\nu\in\cM_+^1(\cX)$.
Typically, the latent space $\cZ$ is of much lower dimensions relative to the data space $\cX$, yielding that both measures $G_\sharp\zeta$ and $\nu$ have singular supports.
Legitimated by the fact that $\cW_c$ metrizes the weak convergence of measures, e.g., if $c(x,y)=\|x-y\|_p^p,~p\geq1$ \cite[Theorem 6.8]{Vil08}, the OT loss $\cW_c$ is well suited as the discrepancy measure between $G_\sharp\zeta$ and $\nu$. The generator learning can be stated as the minimum Kantorovich estimation 
\cite[Section 9.4]{PeCu19}:
\iali{
\inf_{G\in\cC(\cZ;\cX)} \cW_{c}(G_\sharp\zeta,\nu)  &= \inf_{\substack{G\in\cC(\cZ;\cX)\\ \gamma\in\Gamma(G_\sharp\zeta,\nu)}} \int_{\cX\times\cX}c(x,y)\d\gamma(x,y) \label{eq:mke1}\\
&= \inf_{\substack{G\in\cC(\cZ;\cX)\\ \bar\gamma\in\Gamma(\zeta,\nu)}} \int_{\cZ\times\cX}c(G(z),y)\d\bar\gamma(z,y).
\label{eq:mke}
}
The first identity above follows directly from the definition in \eqref{eq:wd}. The second identity was shown in \cite[Theorem 1]{BGTSS17}, which justifies the equivalence of two formulations under the re-parameterization $\gamma=(G,\id)_\sharp\bar\gamma$. This re-parameterization will be further exploited in Section \ref{sec:imse}.

In the literature,
the seminal work on the Wasserstein GANs \cite{ACB17} approaches the training of \eqref{eq:mke1} by introducing discriminator (networks) $D_1,D_2$ into play, i.e.,
\iali{
&\inf_{G\in\cC(\cZ;\cX)} \sup_{D_1,D_2\in\cC(\cX;\bR)}~ \bigg\{ \int_\cZ D_1(G(z))\d\zeta(z) + \int_\cX D_2(y)\d\nu(y) \notag \\
& \qquad  \text{subject to}~ D_1(x)+D_2(y)\leq c(x,y) \text{for }(x,y)~\text{($G_\sharp\zeta\otimes\nu$)-a.e.} \bigg\}.
\label{eq:sdp0}}
The optimal pair $(D_1,D_2)$ complies with the $c$-transform relation \mbox{\cite[Definition 5.2]{Vil08}}:
\iali{
\!\! \forall x\in\cX: D_1(x)=D_2^{(c)}(x):=\inf_{y\in\cX}\big\{c(x,y)-D_2(y)\big\}. \label{eq:ctrans}
}
Provided that $c$ is a metric, the $c$-transform in \eqref{eq:ctrans} specializes to $D_1=-D_2$ with an additional constraint of $D_2$ being 1-Lipschitz, cf.\ \cite[Particular Case 5.4]{Vil08}. This leads to a minimax problem over a generator and a (single) discriminator \cite{ACB17}, akin to the formulation of the original GAN \cite{Goo+14}. The 1-Lipschitz constraint is numerically challenging and often pursued heuristically, e.g., by weight clipping \cite{ACB17}, gradient penalty \cite{GAADC17} or spectral normalization \cite{MKKY18}.

\subsection{Minimum Sinkhorn estimation}
\label{sec:sink}

More recently, the Sinkhorn loss has drawn interests as a smoothed variant for the OT loss; see \cite{GPC18,FSVATP19,GCBCP19}.
With the Kullback-Leibler (KL) divergence defined by 
\iali{
\kld{\mu}{\nu} =
\begin{cases}
\int_\cX\log\left(\frac{\d\mu}{\d\nu}(x)\right) \d\mu(x)\! & \text{if $\frac{\d\mu}{\d\nu}$ exists $\nu$-a.e.} \\
\infty & \text{otherwise,}
\end{cases}
}
the Sinkhorn loss $\cS_{c,\eps}$ is defined through the entropic regularized OT loss $\cW_{c,\eps}$:
\iali{
&\cS_{c,\eps}(\mu,\nu)   = \cW_{c,\eps}(\mu,\nu) - \frac12\cW_{c,\eps}(\mu,\mu) - \frac12\cW_{c,\eps}(\nu,\nu),  \label{eq:skd} \\
&\cW_{c,\eps}(\mu,\nu) = \inf_{\gamma\in\Gamma(\mu,\nu)} \bigg\{ \int_{\cX\times\cX}c(x,y)\d\gamma(x,y) +\eps\cdot\kld{\gamma}{\mu\otimes\nu} \bigg\}. \label{eq:swd}
}
In this work, we consider generative modeling via minimum Sinkhorn estimation:
\iali{
\inf_{G\in\cC(\cZ;\cX)} \cS_{c,\eps}(G_\sharp\zeta,\nu) = \inf_{G\in\cC(\cZ;\cX)} \bigg\{ \cW_{c,\eps}(G_\sharp\zeta,\nu) - \frac12\cW_{c,\eps}(G_\sharp\zeta,G_\sharp\zeta) - \frac12\cW_{c,\eps}(\nu,\nu) \bigg\}.
\label{eq:mse}
}
Promising empirical results on generative modeling in the spirit of \eqref{eq:mse} were obtained recently in \cite{SZRM18,SBRL18}.
Note that the last term $-\frac12\cW_{c,\eps}(\nu,\nu)$ is a constant which can be safely ignored in the training of $G$.

The advantages of using $\cS_{c,\eps}$ rather than $\cW_{c}$ or $\cW_{c,\eps}$ as the training loss are previously observed in literature and summarized in the following:
\bi
\item[(i)] The regularization by KL divergence renders $\cW_{c,\eps}(\mu,\nu)$ (and hence $\cS_{c,\eps}(\mu,\nu)$ as well) Fr\'echet-differentiable with respect to $\mu$, see \cite[Proposition 2]{FSVATP19}. The smoothness of the generative objective was exploited in \cite{SBRL18}.
\item[(ii)] The Sinkhorn loss $\cS_{c,\eps}$ induces a divergence which interpolates between the OT distance $\cW_{c}$ (letting $\eps\to0$) and the maximum mean discrepancy (MMD) \cite{SFGSL12} (letting $\eps\to\infty$). In this respect, $\cS_{c,\eps}$ metrizes the weak convergence of measures same as $\cW_c$ does \cite[Theorem 1]{FSVATP19}, while enjoying a better sample complexity than $\cW_{c,\eps}$ \cite{GPC18,GCBCP19}.
\item[(iii)] The entropic regularization in $\cW_{c,\eps}$ observably introduces bias \cite{SBRL18,FSVATP19} since in general $\cW_{c,\eps}(\mu,\mu)\neq0$.
Remarkably, the inclusion of the \emph{debiasing} term
$-\frac12\cW_{c,\eps}(G_\sharp\zeta,G_\sharp\zeta)$ in \eqref{eq:mse} accounts
for avoiding a mode collapse of $G_\sharp\zeta$ towards a shrunk measure (e.g.,
median of $\nu$ if $c(x,y)=\|x-y\|_1$ or mean of $\nu$ if $c(x,y)=\|x-y\|_2^2$),
see \cite[Figure 1]{FSVATP19} on a 2D toy example and Figure \ref{fig:debias}
on MNIST.
\ei

\begin{figure*}[t!]
  \centering
  \begin{center}
  \begin{subfigure}[t]{0.495\textwidth}
  \centering
    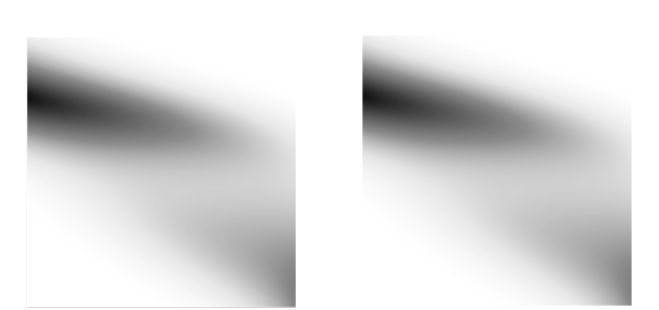
    \caption{$\varepsilon = 20$, $\lambda = 0$}
    \label{fig:plans_a}
  \end{subfigure}
  \begin{subfigure}[t]{0.495\textwidth}
  \centering
    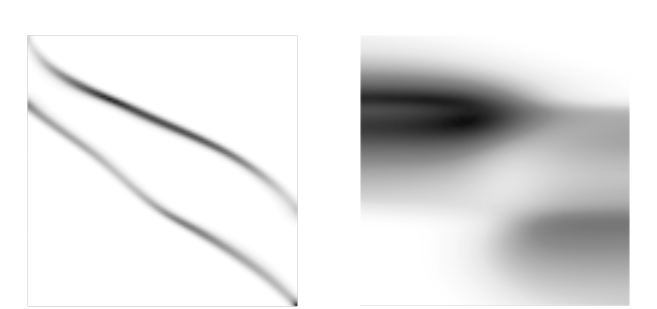
    \caption{$\varepsilon = 20$, $\lambda = 5$}
    \label{fig:plans_b}
  \end{subfigure}
  \end{center}
  \caption{We compute two transportation plans $\gamma'$ and $\gamma$ with $c(x,y)=\norm{x-y}^2$ and marginals $\zeta$ and $\nu$. $\zeta$ is uniform on the unit square, and $\nu$ is a mixture of 1D Gaussians supported only on a diagonal of the unit square.
In (\subref{fig:plans_a}), the plan $\gamma'$ is obtained without informative regularization, i.e.,\ $\lambda=0$ in \eqref{eq:imse1}, and is therefore rather diffuse. For (\subref{fig:plans_b}), we switch on the informative regularization ($\lambda=5$) and observe that it makes the latent dimension $z_1$ \emph{informative}, indicated by a sparse support of $\pi_{(z_1,y)\sharp}\gamma$.
}
  \label{fig:idea}
\end{figure*}

\section{Informative Sinkhorn GANs} \label{sec:imse}
Learning disentangled representations is the holy grail of generative modeling, see the discussion in Section \ref{sec:intro}.
Inspired by mutual information maximization in InfoGAN \cite{CDHSSA16} and InfoVAE \cite{ZSE17}, we propose to include an additive regularization term in the minimum Sinkhorn estimation \eqref{eq:mse} such that a prescribed latent subspace is encouraged to be informative. The key to access this regularization is the implicit transportation plan arising from the smoothed OT loss $\cW_{c,\eps}(G_\sharp\zeta,\nu)$.

\subsection{Implicit transportation plan} \label{sec:htp}
In the context of generative modeling, the Fenchel dual formulation of the smoothed OT loss $\cW_{c,\eps}(G_\sharp\zeta,\nu)$ (recall the definition from \eqref{eq:swd}) is given by
\iali{
 &\cW_{c,\eps}(G_\sharp\zeta,\nu) = \inf_{\gamma\in\Gamma(G_\sharp\zeta,\nu)}
 \bigg\{ \int_{\cX\times\cX}c(x,y)\d\gamma(x,y)
 +\eps\cdot\kld{\gamma}{G_\sharp\zeta\otimes\nu} \bigg\} \label{eq:swgm1} \\
&= \sup_{D_1,D_2\in\cC(\cX;\bR)} \bigg\{ \int_\cZ D_1(G(z))\d\zeta(z) + \int_\cX D_2(y)\d\nu(y) \notag \\ 
& \qquad \qquad -\int_{\cZ\times\cX} \eps\exp\Big( \frac1\eps\big( D_1(G(z))+D_2(y) -c(G(z),y) \big) \Big) \d\zeta\otimes\nu(z,y) \bigg\}.
}
The optimal pair $\gamma$ and $(D_1, D_2)$ satisfies the conditions:
\iali{
& \pi_{x\sharp}\gamma = G_\sharp\zeta,~~\pi_{y\sharp}\gamma=\nu, \\
& \frac{\d\gamma}{d (G_\sharp\zeta\otimes\nu)}(x,y) = \exp\Big( \frac1\eps\big( D_1(x)+D_2(y)-c(x,y) \big) \Big). \label{eq:swoc2}
}
In particular, these conditions guarantee that it is viable to re-parameterize $\gamma=(G,\id)_\sharp\bar\gamma$ as stated in Theorem \ref{thm:reparam}.
\bthm \label{thm:reparam}
It is equivalent to re-parameterize $\gamma=(G,\id)_\sharp\bar\gamma$ for the infimum in \eqref{eq:swgm1}, that is
\iali{
&\cW_{c,\eps}(G_\sharp\zeta,\nu)
= \inf_{\bar\gamma\in\Gamma(\zeta,\nu)} \bigg\{
\int_{\cZ\times\cX}c(G(z),y)\d\bar\gamma(z,y)
+\eps\cdot\kld{(G,\id)_\sharp\bar\gamma}{G_\sharp\zeta\otimes\nu} \bigg\}.
\label{eq:swgm2}
}
\ethm
\bpf
On the one hand, $\gamma=(G,\id)_\sharp\bar\gamma$ for $\bar\gamma\in\Gamma(\zeta,\nu)$ ensures the feasibility that $\gamma\in\Gamma(G_\sharp\zeta,\nu)$, and this specialization of $\gamma$ yields an inequality of \eqref{eq:swgm2} in the direction ``$\leq$''.
On the other hand, the optimal $\gamma$ satisfying \eqref{eq:swoc2} takes the form $\gamma=(G,\id)_\sharp\bar\gamma$ with
\iali{
\frac{\d\bar\gamma}{d (\zeta\otimes\nu)}(z,y)  = \exp\Big( \frac1\eps\big( D_1(G(z))+D_2(y)-c(G(z),y) \big) \Big),
}
and hence \eqref{eq:swgm2} indeed holds with an equality.
\epf

This implicit transportation plan $\bar\gamma\in\cM_+^1(\cZ\times\cX)$ contains rich information as $\bar\gamma$ is a reasonable proxy for the joint distribution of the latent code and the generated image.
We will further exploit this observation in the next subsection so as to make generator $G$ \emph{informative}.

\subsection{Informative minimum Sinkhorn estimation via structured regularization} \label{sec:infosink}
We first introduce necessary notations. Let the latent space be decomposed as $\cZ=\cZ_1\times\cZ_2$ with the intended informative subspace $\cZ_1$ and its noisy counterpart $\cZ_2$. Furthermore, let $\pi_{z_1}(z_1,z_2)=z_1,~\pi_{(z_1,y)}(z_1,z_2,y)=(z_1,y)$ and $\zeta=\zeta_1\otimes\zeta_2$ with $\zeta_1\in\cM_+^1(\cZ_1)$ and $\zeta_2\in\cM_+^1(\cZ_2)$.

Given a latent code $z=(z_1,z_2)\sim\zeta$, $z_1$ being informative would yield high \emph{mutual information} between $z_1$ and $x=G(z)$, which can be quantified through the KL divergence \cite[Section 2.3]{CoTh06}:
\iali{
&\MI((z_1,x)\sim(\pi_{z_1},G)_\sharp\zeta) 
= \kld{(\pi_{z_1},G)_\sharp\zeta}{\zeta_1\otimes G_\sharp\zeta}.
\label{eq:mi}
}
Invoking the transportation plan $\bar\gamma$ in the formulation \eqref{eq:swgm2} as a proxy for $(\pi_{z_1},G)_\sharp\zeta$, we come up with the following \emph{informative minimum Sinkhorn estimation}:
\iali{
& \inf_{\substack{G\in\cC(\cZ;\cX)\\ \bar\gamma\in\Gamma(\zeta,\nu)}}
\bigg\{ \int_{\cZ\times\cX}c(G(z),y)\d\bar\gamma(z,y)
+\eps\cdot\kld{(G,\id)_\sharp\bar\gamma}{G_\sharp\zeta\otimes\nu} \notag \\
& \qquad \qquad \qquad -\frac12\cW_{c,\eps}(G_\sharp\zeta,G_\sharp\zeta)
-\lambda\cdot\kld{\pi_{(z_1,y)\sharp}\bar\gamma}{\zeta_1\otimes\nu} \bigg\}.
\label{eq:imse1}
}

The structured entropic regularization (the negative $\lambda\cdot$KL term) drives entropy of the marginalized transportation plan $\pi_{(z_1,y)\sharp}\bar\gamma$ low and in return $z_1\sim\zeta_1$ to be informative.
Figure \ref{fig:idea} illustrates the effect of this regularization through a toy example. For that example we fix $G=\id$ and for the numerical solution, we adopt a DC
  strategy~\cite{horst1999dc} and solve a sequence of perturbed problems with the Sinkhorn algorithm \cite{Cut13}.

To gain further insights on our model \eqref{eq:imse1}, let us consider the dual formulation of the KL divergence used in
\cite{nguyen2010estimating,nowozin2016f,belghazi2018mine}:
\iali{
\kld{\pi_{(z_1,y)\sharp}\bar\gamma}{\zeta_1\otimes\nu}  &= \sup_{Q\in\cC(\cZ_1\times\cX;\bR)} \bigg\{ \int_{\cZ\times\cX} Q(z_1,y)\d\bar\gamma(z,y) +1 \notag \\
                                                        &\qquad -\int_{\cZ_1\times\cX} \exp\big( Q(z_1,y)\big)\d\zeta_1\otimes\nu(z_1,y) \bigg\}.
\label{eq:bound}
}
For a formal proof of the above equality, we refer the interested reader to \cite[Proposition 7]{FSVATP19}. Using the dual formulation \eqref{eq:bound}, we convert \eqref{eq:imse1} into the form:
\iali{
&\inf_{\substack{G\in\cC(\cZ;\cX)\\ \bar\gamma\in\Gamma(\zeta,\nu)\\ Q\in\cC(\cZ_1\times\cX;\bR)}} \bigg\{ \int_{\cZ\times\cX}\Big(c(G(z),y) - \lambda Q(z_1,y)\Big)\d\bar\gamma(z,y)
+\eps\cdot\kld{(G,\id)_\sharp\bar\gamma}{G_\sharp\zeta\otimes\nu} \notag \\
& \qquad \qquad \qquad -\frac12\cW_{c,\eps}(G_\sharp\zeta,G_\sharp\zeta) \lambda +\lambda\int_{\cZ_1\times\cX} \exp\big( Q(z_1,y)\big)\d\zeta_1\otimes\nu(z_1,y)  \bigg\}.
\label{eq:imse2}
}
That way,
we can interpret the variational variable $Q$ in \eqref{eq:imse2} in the language of optimal transport:
$Q$ learns to modify the (pulled-back) ground cost $c\circ(G,\id)$,
with a bias towards making the marginalized transportation plan $\pi_{(z_1,y)\sharp}\bar\gamma$ informative (i.e.\ low entropy).

Note that there are various ways to obtain variational lower bound to the mutual information, see \cite{poole2018variational} for a recent survey. Our bound \eqref{eq:bound}
is based on convex duality, which we found to be natural within the framework of optimal transport. Another bound is
the one of Barber and Agakov \cite{barber2003algorithm} given by
\begin{equation}
  \kld{\pi_{(z_1,y)\sharp}\bar\gamma}{\zeta_1\otimes\nu} \geq \int_{\cZ \times \cX} \log q(z_1 | y) \, \d\bar\gamma(z,y),
    \label{eq:ba}
\end{equation}
where $q( z_1 | y)$ is the density, with respect to $\zeta_1$, of an arbitrary conditional distribution on $z_1$ given $y$.
Note that \eqref{eq:ba} can be recovered from \eqref{eq:bound}
by selecting the special parametrization $Q(z_1, y) = \log q(z_1 | y)$.

The main conceptual difference to InfoGAN \cite{CDHSSA16}, which maximizes the mutual information between the latent
and generated variables \eqref{eq:mi} via the variational lower bound \eqref{eq:ba}, is that we consider the transport
plan $\bar \gamma$ as a proxy for the true, but intractable, joint distribution.  Interestingly enough, it can be noted that
our $Q$ only operates on real data samples, as opposed to the ``$Q$ network'' developed in
\cite{CDHSSA16}. In such a setting, our $Q$ influences the generator $G$ through modifying the (pulled back) ground cost.

\subsection{A practical training algorithm} \label{sec:train}
Here we derive a practical training scheme for model \eqref{eq:imse1}. Based on the formulation \eqref{eq:imse2},
we compute the associated OT loss via its dual formulation:
{
\allowdisplaybreaks
\iali{
&\inf_{\bar\gamma\in\Gamma(\zeta,\nu)} \bigg\{ \int_{\cZ\times\cX}\Big(c(G(z),y) - \lambda Q(z_1,y)\Big)\d\bar\gamma(z,y) 
+\eps\cdot\kld{(G,\id)_\sharp\bar\gamma}{G_\sharp\zeta\otimes\nu} \bigg\} \notag \\
& = \sup_{D_1,D_2\in\cC(\cX;\bR)} \bigg\{ \int_{\cZ} D_1(G(z))\d\zeta(z) +
\int_{\cX} D_2(y)\d\nu(y) +\eps \notag \\ 
& \qquad \qquad  -\eps\int_{\cZ\times\cX} \exp\Big(
\frac1\eps\big( D_1(G(z))+D_2(y) +\lambda Q(z_1,y)-c(G(z),y) \big) \Big)
\d\zeta\otimes\nu(z,y) \bigg\}, \label{eq:wzyd-1}\\
&= \sup_{D_2\in\cC(\cX;\bR)} \bigg\{ \int_{\cZ} D_2^{(c,\eps)}(G(z))\d\zeta(z) + \int_{\cX} D_2(y)\d\nu(y) +\eps \bigg\}, \label{eq:wzyd}
}
where $D_1$ is substituted by the $(c,\eps)$-transform \cite[Section 5.3]{PeCu19} of $D_2$:
\iali{
&D_2^{(c,\eps)}(\cdot)
= -\eps\log\int_\cX\exp\Big(\frac1\eps\big(D_2(y)-c(\cdot,y)\big) \Big) \d\nu(y).
}
Compared to the $c$-transform in \eqref{eq:ctrans}, the $(c,\eps)$-transform can be viewed as a soft pointwise minimum on $c(\cdot,y)-D_2(y)$ over $y$.
The derivation for \eqref{eq:wzyd-1}--\eqref{eq:wzyd} follows from an application of the Fenchel-Rockafellar’s theorem and can be found in \cite[Proposition 2.1]{GCPB16}.

In a similar way, the second OT loss $\cW_{c,\eps}(G_\sharp\zeta,G_\sharp\zeta)$ in \eqref{eq:imse2} is computed as
\iali{
&\cW_{c,\eps}(G_\sharp\zeta,G_\sharp\zeta) \notag\\&
= \inf_{\bar\eta\in\Gamma(\zeta,\zeta)} \bigg\{ \int_{\cZ\times\cZ}
c(G(z'),G(z'')) \d\bar\eta(z',z'')
+\eps\cdot\kld{(G,G)_\sharp\bar\eta}{G_\sharp\zeta\otimes G_\sharp\zeta}
\bigg\}
\\ &
= \sup_{D_4\in\cC(\cX;\bR)} \bigg\{ \int_\cZ D_4^{(c,\eps)}(G(z'))\d\zeta(z') 
+ \int_\cZ D_4(G(z''))\d\zeta(z'') +\eps \bigg\}.
\label{eq:wzzd}
}
Once again, a dual variable $D_3$ has been eliminated via the $(c,\eps)$-transform:
\iali{
& D_3(\cdot) =D_4^{(c,\eps)}(\cdot) \notag\\& = -\eps\log\int_\cZ\exp\Big(\frac1\eps\big(D_4(G(z'')) 
-c(\cdot,G(z''))\big) \Big) \d\zeta(z'').
}

With optimal dual variables $D_2$ and $D_4$ in hand,
the optimal primal variables $\bar\gamma$ and $\bar\eta$ can be retrieved from the optimality conditions:
\iali{
\frac{\d\bar\gamma}{d(\zeta\otimes\nu)}(z,y) =\,& \exp\Big( \frac1\eps\big(
D_2^{(c,\eps)}(G(z))+D_2(y) +\lambda Q(z_1,y)-c(G(z),y) \big)\Big),
\label{eq:wzyo}\\
\frac{\d\bar\eta}{d(\zeta\otimes\zeta)}(z',z'') =\,& \exp\Big(
\frac1\eps\big(D_4^{(c,\eps)}(G(z'))+D_4(G(z''))  -c(G(z'),G(z'')) \big)\Big).
\label{eq:wzzo}
}
Our training algorithm alternates between $(\bar\gamma,\bar\eta)$ and $(Q,G)$.
Specifically, we use the supremum problem \eqref{eq:wzyd} over $D_2$ (resp.\ \eqref{eq:wzzd} over $D_4$) and relation \eqref{eq:wzyo} (resp.\ \eqref{eq:wzzo}) as an oracle to update the expression for $\bar\gamma$ (resp.\ $\bar\eta$).
Alternatively, given $\bar\gamma$ and $\bar\eta$ the variables $Q$ and $G$ can be updated in parallel through the following optimization:
\iali{
& \inf_{Q\in\cC(\cZ_1\times\cX;\bR)} \int_{\cZ\times\cX} - Q(z_1,y)\d\bar\gamma(z,y)  +\int_{\cZ_1\times\cX} \exp\big( Q(z_1,y)\big)\d\zeta_1\otimes\nu(z_1,y), \label{eq:qsub} \\
& \inf_{G\in\cC(\cZ;\cX)} \int_{\cZ\times\cX} c(G(z),y)\d\bar\gamma(z,y)  -\frac12\int_{\cZ\times\cZ}c(G(z'),G(z''))\d\bar\eta(z',z''). \label{eq:gsub}
}
The overall algorithm is spelled out in Algorithm~\ref{alg}.
}

\balg[t]
\caption{Training scheme for model \eqref{eq:imse1}.}
\begin{algorithmic}[1]
\label{alg}
\REQUIRE{$\zeta\in\cM_+^1(\cZ),~\nu\in\cM_+^1(\cX),~c\in\cC(\cX\times\cX;\bR_+),~\eps,\lambda>0,~n_D\in\bN$.}
\STATE{Initialize $D_2,D_4\in\cC(\cX;\bR),\, G\in\cC(\cZ;\cX),\, Q\in\cC(\cZ_1\times\cX;\bR)$ as neural networks.
}
\WHILE{$G$ not converged}
\FOR{$t=1,...,n_D$}
\STATE{Draw minibatch samples $(z,y)\sim\zeta\otimes\nu$. }
\STATE{Update the weights of $D_2$ by an Adam step on \eqref{eq:wzyd}.}
\STATE{Draw minibatch samples $(z',z'')\sim\zeta\otimes\zeta$. }
\STATE{Update the weights of $D_4$ by an Adam step on \eqref{eq:wzzd}.}
\ENDFOR
\STATE{Express $\bar\gamma\in\cM_+^1(\cZ\times\cX)$ using \eqref{eq:wzyo} and $\bar\eta\in\cM_+^1(\cZ\times\cZ)$ using \eqref{eq:wzzo}.}
\STATE{Draw minibatch samples $(z,y,z',z'')\sim\zeta\otimes\nu\otimes\zeta\otimes\zeta$.}
\STATE{Update in parallel the weights of $Q$ by an Adam step on \eqref{eq:qsub} and the weights of $G$ by an Adam step on \eqref{eq:gsub}.}
\ENDWHILE
\end{algorithmic}
\ealg

    \begin{figure*}[t!]
        \centering
        \begin{subfigure}[t]{0.249\textwidth}
            \includegraphics[width=\linewidth,max height=1.00\textheight]{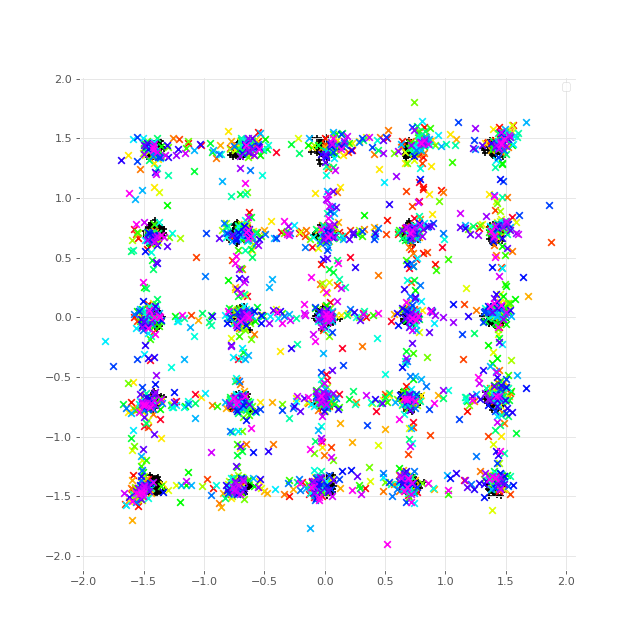}
            \caption{Sinkhorn}
            \label{fig:vary-cat_gaussian_a}
        \end{subfigure}
        \begin{subfigure}[t]{0.249\textwidth}
            \includegraphics[width=\linewidth,max height=1.00\textheight]{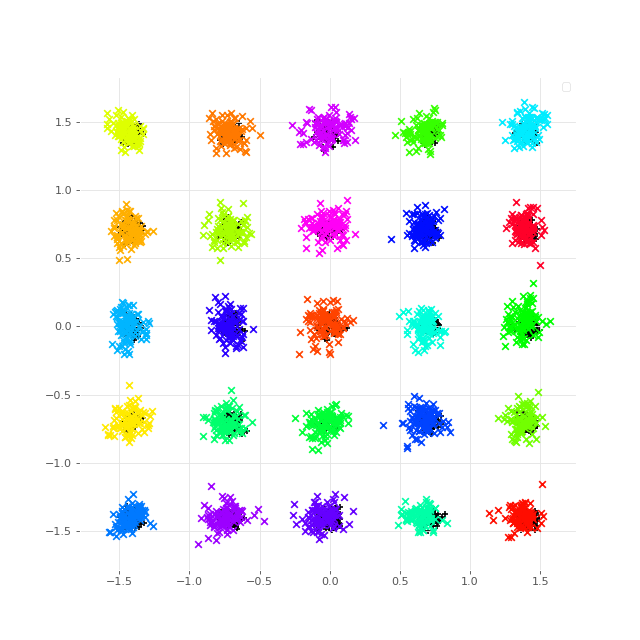}
            \caption{Info-Sinkhorn}
            \label{fig:vary-cat_gaussian_b}
        \end{subfigure}
   \centering
\begin{subfigure}[t]{0.24\textwidth}
   \includegraphics[width=\linewidth,max height=1.00\textheight]{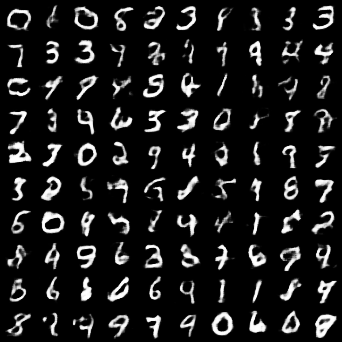}
   \caption{Sinkhorn}
   \label{fig:vary-cat_mnist_a}
\end{subfigure}
\begin{subfigure}[t]{0.24\textwidth}
   \includegraphics[width=\linewidth,max height=1.00\textheight]{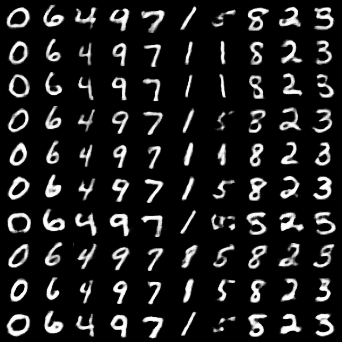}
   \caption{Info-Sinkhorn}
   \label{fig:vary-cat_mnist_b}
\end{subfigure}
\caption{Variation of the categorical latent code on Gaussian dataset and MNIST, using $c(x, y) =
\|x-y\|_1$, $\eps = \lambda = 0.05$. The categorical code is color-coded for
the Gaussian dataset. From MNIST, each line has its categorical encoding
varied while its other latent dimensions are kept constant. The continuous
informative codes are set to zero, and for each row, the noisy dimensions are
drawn according to $\zeta_2$. This categorical encoding is clear only for
the informative models (\subref{fig:vary-cat_mnist_b}).}
   \label{fig:vary-cat_mnist}
\end{figure*}

\section{Experiments}
\label{sec:exp}

In this section, we demonstrate and
discuss experimental results for the Info-Sinkhorn model trained by
Algorithm~\ref{alg}. Our results are compared qualitatively to the baseline Sinkhorn model over two datasets: 2D multimodal Gaussians and MNIST. 

\subsection{Experimental setup}

\paragraph{Regularization parameters.} Two main hyper-parameters are critical in the
algorithm: $\eps$ and $\lambda$. A good balanced choice is to have $\lambda$ equal to or 
slightly less than $\eps$. Typically, we set $\lambda = \eps \approx 0.05$. 
The complete training details can be found in Appendix~\ref{app:training-details}.

\paragraph{Network architectures.} The network architectures are adapted
from previous works \cite{CDHSSA16,SBRL18,GAADC17}.

In particular, the discriminators $D_{\{2, 4\}}, D_Q$ are set to share their first
(convolutional) layers and only learn the last layers independently. $D_Q$ shares
its first layers with the discriminators $D_{\{2, 4\}}$ as in \cite{CDHSSA16}.

The update of the shared parameters is only performed during the update of the
dual networks $D_{\{2,4\}}$.
Extensive sharing of layers and weights among the networks $D_{\{2, 4\}}$ and
$D_Q$ yields minimal computational overhead of the proposed informative
generative model compared to existing GANs \cite{Goo+14,ACB17,CDHSSA16}.

\paragraph{Parametrization of $Q$.} The Info-Sinkhorn model offers the liberty to parametrize the function $Q$ that appears in \eqref{eq:imse2}. In practice, we take a simple choice of a dot product rule, namely $Q(z_1, x) := z_1 ^\top
D_Q(x)$ where a network $D_Q$ outputs a vector of the same size as $z_1$. This parametrization, though being linear in $z_1$, already yields good
disentanglement results on MNIST (see Section \ref{sec:mnist}), and empirically works better than the $\log$-parametrization from \eqref{eq:ba} aligned with the InfoGAN \cite{CDHSSA16}. 

\begin{figure*}[!t] \centering
    \begin{subfigure}[t]{0.32\textwidth}
    \includegraphics[width=\linewidth,max height=0.50\textheight]{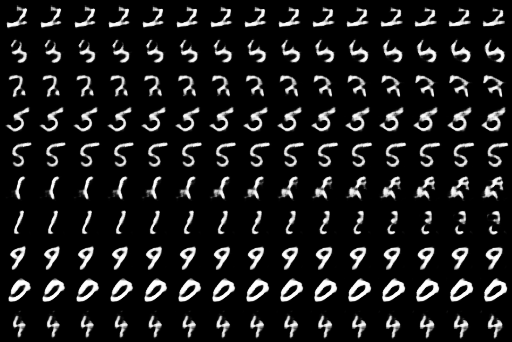}
    \caption{Sinkhorn, $z_2$} \label{fig:vary-info_mnist_a} \end{subfigure}
    \begin{subfigure}[t]{0.32\textwidth}
        \includegraphics[width=\linewidth,max height=0.50\textheight]{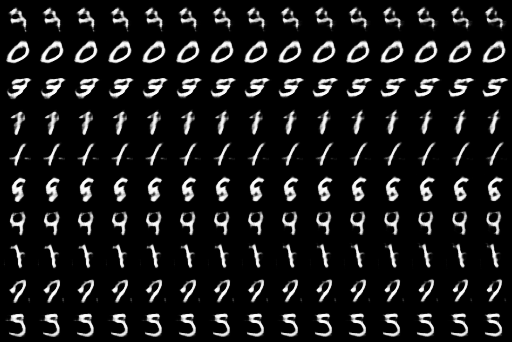}
        \caption{Sinkhorn, $z_3$} \label{fig:vary-info_mnist_b} \end{subfigure}
        \begin{subfigure}[t]{0.32\textwidth}
        \includegraphics[width=\linewidth,max height=0.50\textheight]{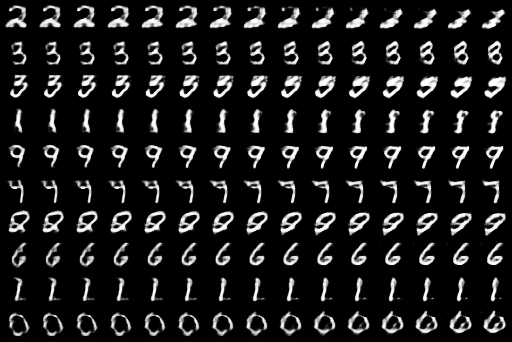}
\caption{Sinkhorn, $z_4$\vspace{0.2cm}} \label{fig:vary-info_mnist_c}
\end{subfigure}

\begin{subfigure}[t]{0.32\textwidth}
    \includegraphics[width=\linewidth,max height=0.50\textheight]{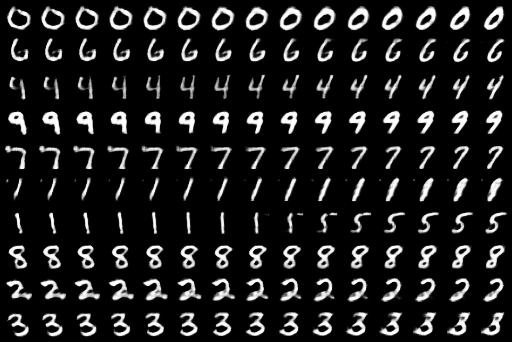}
    \caption{Info-Sinkhorn, $z_2$} \label{fig:vary-info_mnist_d}
    \end{subfigure} \begin{subfigure}[t]{0.32\textwidth}
    \includegraphics[width=\linewidth,max height=0.50\textheight]{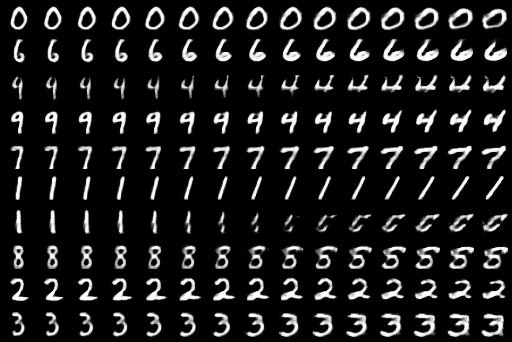}
    \caption{Info-Sinkhorn, $z_3$} \label{fig:vary-info_mnist_e}
    \end{subfigure} \begin{subfigure}[t]{0.32\textwidth}
\includegraphics[width=\linewidth,max height=0.50\textheight]{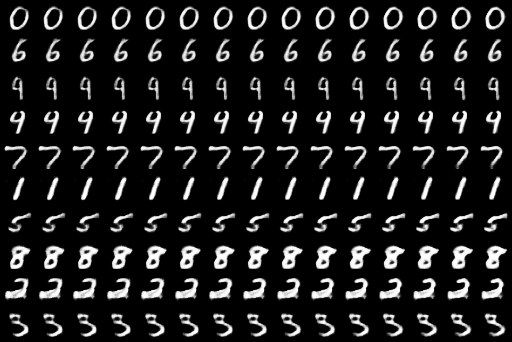}
\caption{Info-Sinkhorn, $z_4$} \label{fig:vary-info_mnist_f} \end{subfigure}

\caption{Variations in the latent space on the continuous code for MNIST. Upper
    row: Sinkhorn. Lower row: Info-Sinkhorn. $\eps = \lambda = 0.05$. Left two columns: two continuous
    informative latent variables $z_2, z_3$. Right column: first noisy latent
    variable $z_4$. In each picture, each line represents a draw in the sample
    space, with a different categorical code, and which is fixed across all
    columns \emph{except} for the appropriate continuous code $z_i$, which
    varies between $-1$ and $1$. There is a significant and interpretable
    visual variation only for the informative dimensions in the Info-Sinkhorn
    model (\subref{fig:vary-info_mnist_d}),(\subref{fig:vary-info_mnist_e}).}
\label{fig:vary-info_mnist} \end{figure*}

    \subsection{2D multimodal Gaussian dataset}
    A toy dataset is considered for testing the informative model, as it
    illustrates the learning algorithm quite well. A 25-modal 2D Gaussian
    dataset is learned, with a latent space factored into a 25-dimensional
    categorical (informative) part, and a 103 dimensional Gaussian (noisy)
    part.  As can been seen in \cref{fig:vary-cat_gaussian_a}, the different
    modes are perfectly recovered, and even less artefacts are observed (as
    opposed to the baseline Sinkhorn model).

    Although structurally simple, this dataset can be proven difficult to learn
    for standard GANs as mentioned in \cite{SBRL18}. This illustrates the effectiveness of OT-based
    generative models when the underlying Euclidean space metric makes sense
    for comparing the samples: here the cost between samples $c(x, y) = \| x -
    y \|$ is natural. This strongly suggests that the proposed
    informative regularization on the transport plan can be relevant.

\subsection{Latent space traversal on MNIST}
\label{sec:mnist}
We further test Info-Sinkhorn on MNIST.
The latent factorization in this case is taken as follows: $z_1$ is an
informative categorical code of dimension $10$, $z_2, z_3\sim\text{Unif}[-1,1]$
are informative continuous latent codes. The remaining latent space
$(z_j)_{j=4}^{91}\sim \cN(0,1)$ is designated to be the uninformative
dimension. 

In the different visualizations, each row represent a latent space traversal,
and therefore has only one latent code that varies along the columns, whereas
the other dimensions remain identical. This allows to visually evaluate the
disentanglement and interpretability of the latent representation.

In \cref{fig:vary-cat_mnist_b}, the ten categories in MNIST are
 well learned with the Info-Sinkhorn model. In
contrast, for the baseline Sinkhorn model shown in
\cref{fig:vary-cat_mnist_a} the categorical latent code has no
clear interpretation. Moreover, given a row, the digit aspect remains
similar across each columns. This shows how the style of a
sample is mostly determined by the continuous latent space.

This is presented in more detail in \cref{fig:vary-info_mnist}.
The informative model is compared to the Sinkhorn one, as to
illustrate the effect of the proposed regularization on the
output of the network. Although the visual cues are not
perfectly disentangled (e.g.\ the rotation of certain digits
is a result of both variations), the contrast is clear with
the original Sinkhorn model, where the latent space
configuration has unpredictable effects on the visual
samples. The informative regularization yields on the
contrary a much more disentangled and informative latent
space.

\section{Conclusion}
\label{sec:disc}
This work advocates informative generative modeling via optimal transport. The entropic smoothing on an optimal transport loss enables access to the transportation plan. In the context of generative modeling, we introduce a novel structured regularization on this plan to make the prescribed latent dimensions informative.
In this sense, our work bridges a gap between previous informative GANs \cite{CDHSSA16} and smoothed optimal transport \cite{PeCu19}. 

Practically, we derive an efficient training scheme for the proposed informative generative model, which is further boosted by extensive sharing of architectures and weights among multiple ``discriminator'' networks.
We experimentally confirm that the recently introduced Sinkhorn loss \cite{GPC18,FSVATP19} indeed avoids collapses and
that our proposed regularization yields informative latent representation and improved sample quality.

Although limited to a simple dataset, those experiments are
a good sign of possible extensions for optimal transport GANs based on
a structured transport plan regularization. The results in an Euclidean space where the
ground cost was natural to compare samples suggests the offered method has potentials.

The generalization to more complex datasets is subject to
defining/learning a better ground cost function $c(x,y)$, as optimal
transport relies exhaustively on the metric of the
underlying Euclidean space. For complex images, the
pixel-wise metrics are not particularly adequate, and tend
to produce blurred samples with very low variability. One
dominant axis of research should be dedicated to finding a
stable way to learn the cost jointly with the informative
regularization.

\small
\bibliographystyle{plain}
\bibliography{reference}

\clearpage
\newpage
\begin{appendices}

    \section{Training details}
    \label{app:training-details}
    In the reported experiments, the Info-Sinkhorn model was trained using the following hyper-parameters. The
    baseline Sinkhorn model was trained with the exact same set of parameters, except for
    $\lambda=0$.

    \subsection{Gaussians}

    \begin{table}[h]
        \centering
        \begin{tabular}[hb]{|c|c|}
            \hline
            ground cost & $c(x,y) = \|x-y\|_1$\\ \hline
            $d_{\text{cat}}$ & 25\\ \hline
            $d_{\text{uni}}$ & 0\\ \hline
            $d_{\text{noise}}$ & 103\\ \hline
            $\eps$ & 1.5\\ \hline
            $\lambda$ & 1.0\\ \hline
            $n_D$ & 5\\ \hline
            Learning rate for $D_i$ & 0.0002\\ \hline
            Learning rate for $G$ & 0.0002\\ \hline
            Number of generator iter. & 50000\\ \hline
            Adam parameters & $\beta_1=0,\,\beta_2=0.9$\\ \hline
        \end{tabular}
        \caption{Hyper-parameters for the Gaussian dataset}
        \label{tab:pp_pp_compare_gaussian-zt-128_swgan-l1-info-sinkhorn_rotgan}
    \end{table}

    \subsection{MNIST}

    \begin{table}[h]
        \centering
        \begin{tabular}[hb]{|c|c|}
            \hline
            ground cost & $c(x, y) = \|x-y\|_1$\\ \hline
            $d_{\text{cat}}$ & 10\\ \hline
            $d_{\text{uni}}$ & 2\\ \hline
            $d_{\text{noise}}$ & 88\\ \hline
            $\eps$ & 0.05\\ \hline
            $\lambda$ & 0.05 \\ \hline
            $n_D$ & 5\\ \hline
            Weight decay on $Q$ & 0.001 \\ \hline
            Learning rate for $D_i, Q$ & 0.0002\\ \hline
            Learning rate for $G$ & 0.0002\\ \hline
            Number of generator iter. & 24200\\ \hline
            Adam parameters& $\beta_1=0.5,\,\beta_2=0.99$\\ \hline
        \end{tabular}
        \caption{Hyper-parameters for the MNIST dataset}
        \label{tab:pp_pp_compare_mnist-20-bn-info-sinkhorn_reg-entropy_rotgan}
    \end{table}

    \begin{table}[!h]
        \begin{subtable}{\columnwidth}
            \centering
            \begin{tabular}[t]{|c|l|}
                \hline
                \multicolumn{2}{|l|}{\textbf{Networks} $D_{\{2,4\}},\,D_{Q}$}\\  \hline
                \multirow{3}{*}{shared} &{Input 2D point}\\ \cline{2-2}
                                        &{FC 512 -- b.n. --  LReLU} \\ \cline{2-2}
                                        &{FC 512 -- b.n. -- LReLU }\\
                                        \hline
                \multirow{1}{*}{spec} &FC 512 -- b.n. -- LReLU -- $\big[\begin{array}{c}
                        \text{FC 1 for} D_{\{2, 4\}} \\
                        \text{FC } |\cZ_1| \text{ for } D_Q
                    \end{array}
                    \big.$\\ 
                \hline
                \hline
                        \multicolumn{2}{|l|}{\textbf{Generator} $G$}\\ \hline
                        \multicolumn{2}{|l|}{Input $z \in \cZ \subset \bR^{128}$}\\ \hline
                        \multicolumn{2}{|l|}{FC 512 -- LReLU}\\  \hline
                        \multicolumn{2}{|l|}{FC 512 -- LReLU }\\ \hline
                        \multicolumn{2}{|l|}{FC 512 -- LReLU -- FC 2}\\ \hline

            \end{tabular}
            \label{tab:arch-gaussian}
            \caption{ Architecture for generating 2D multi-modal Gaussian data; adapted from \cite{GAADC17}.}
        \end{subtable}
        \vspace{\baselineskip}
        \begin{subtable}{\columnwidth}
            \centering
            \begin{tabular}[t]{|c|l|}
                \hline
                \multicolumn{2}{|l|}{\textbf{Networks} $D_{\{2,4\}},\,D_Q$} \\ \hline
                \multirow{4}{*}{shared}  &{Input 32x32x1 Gray image pixels} \\
                \cline{2-2}
                                         &{4 x 4 conv. 64 (stride 2) -- b.n. -- LReLU} \\
                                         \cline{2-2}
                                         &{4 x 4 conv. 128 (stride 2) -- b.n. -- LReLU }\\
                                         \cline{2-2}
                                         & {FC 1024 -- b.n. -- LReLU} \\
                                         \hline
                \multirow{1}{*}{spec} & -- $\big[\begin{array}{c}
                        \text{FC 1 for } D_{\{2,4\}} \\
                        \text{FC 128 -- b.n. -- LReLU -- FC } |\cZ_1| \text{ for } D_Q
                    \end{array}\big.$  \\
                \hline
                \hline
                \multicolumn{2}{|l|}{\textbf{Generator} $G$}\\ \hline
                \multicolumn{2}{|l|}{Input $z \in \cZ \subset \bR^{100}$}\\ \hline
                \multicolumn{2}{|l|}{FC 1024 -- b.n. -- LReLU} \\ \hline 
                \multicolumn{2}{|l|}{FC $8\cdot8\cdot128$ -- b.n. -- LReLU }\\ \hline 
                \multicolumn{2}{|l|}{4 x 4 transp. conv. 64 (stride 2) -- b.n. -- LReLU}\\ \hline
                \multicolumn{2}{|l|}{4 x 4 transp. conv. 1 (stride 2) -- tanh }\\ \hline
            \end{tabular}
            \label{tab:arch-mnist}
            \caption{Architecture for generating MNIST; adapted from \cite{CDHSSA16}.}
        \end{subtable}
        \caption{Architectures for the two datasets. Output dimension for fully
            connected layers (FC) is indicated. The kernel sizes and strides for 2D convolution
            (conv.) and transpose convolution (transp. conv.) are indicated, the
            input and output shapes being given by spatial convolution artihmetics.
            LeakyReLU (LReLU) activation functions with negative slope of $0.2$ are used in
        both generator and dual networks. Batchnorm (b.n.) is used when indicated.}
        \label{tab:architectures}
    \end{table}

\end{appendices}

\end{document}